\documentclass[conference]{IEEEtran}

\IEEEoverridecommandlockouts
\usepackage{cite}
\usepackage{amsmath,amssymb,amsfonts}
\usepackage{algorithmic}
\usepackage{graphicx}
\usepackage{textcomp}
\usepackage{xcolor}

\usepackage{multirow}
\usepackage{booktabs}
\usepackage[ruled,linesnumbered]{algorithm2e}
\usepackage[normalem]{ulem}
\usepackage{balance}
\usepackage{threeparttable}
\usepackage{url}

\def\BibTeX{{\rm B\kern-.05em{\sc i\kern-.025em b}\kern-.08em
    T\kern-.1667em\lower.7ex\hbox{E}\kern-.125emX}}
\begin{document}

\title{DeepGate: Learning Neural Representations of Logic Gates}

\author{\IEEEauthorblockN{Min Li\IEEEauthorrefmark{1}, Sadaf Khan\IEEEauthorrefmark{1}, Zhengyuan Shi\IEEEauthorrefmark{1}, Naixing Wang\IEEEauthorrefmark{2}, Yu Huang\IEEEauthorrefmark{2} and Qiang Xu\IEEEauthorrefmark{1}}\\
\IEEEauthorblockA{\IEEEauthorrefmark{1}
\textit{The Chinese University of Hong Kong},
Shatin, Hong Kong S.A.R.\\
\IEEEauthorblockA{\IEEEauthorrefmark{2}\textit{Huawei Technologies Co., Ltd.}, China \\
\{mli, skhan, zyshi21, qxu\}@cse.cuhk.edu.hk}
}
}

\maketitle


\begin{abstract}
Applying deep learning (DL) techniques in the electronic design automation (EDA) field has become a trending topic. Most solutions apply well-developed DL models to solve specific EDA problems. While demonstrating promising results, they require careful model tuning for every problem. The fundamental question on \textit{"How to obtain a general and effective neural representation of circuits?"} has not been answered yet. In this work, we take the first step towards solving this problem. We propose \textit{DeepGate}, a novel representation learning solution that effectively embeds both logic function and structural information of a circuit as vectors on each gate. Specifically, we propose transforming circuits into unified and-inverter graph format for learning and using signal probabilities as the supervision task in DeepGate. We then introduce a novel graph neural network that uses strong inductive biases in practical circuits as learning priors for signal probability prediction. Our experimental results show the efficacy and generalization capability of DeepGate. 

\end{abstract}

\vspace{5pt}
\section{Introduction}
\label{sec:intro}
\vspace{5pt}

The rise of deep learning (DL) has aroused much interest in applying it to solve various electronic design automation (EDA) problems that otherwise rely on some heuristics or hand-engineered rules, as surveyed in~\cite{huang2021machine}. The most natural representation of circuits and netlists is a graph. With the recent success of graph neural networks (GNNs)~\cite{kipf2016semi,hamilton2017inductive} in modeling non-structured data, various works have explored its potential on EDA problems such as congestion prediction~\cite{kirby2019congestionnet} and testability analysis~\cite{ma2019high}. These works focus on learning a particular function that takes the circuit graph as input and directly maps it to output for desired EDA tasks, without considering the internal computational process in the circuits. 

Recently, a notable trend in the deep learning community is to employ pre-trained models for many downstream tasks rather than learning a specific model for each task from scratch~\cite{han2021pre}. For example, a series of convolutional neural networks (CNNs) are pre-trained on the ImageNet dataset~\cite{krizhevsky2012imagenet}. They perform well on other computer vision (CV) tasks such as image segmentation~\cite{minaee2021image} and object detection~\cite{ren2015faster} by fine-tuning with a small amount of task-specific data. Similarly, pre-trained Transformer-based language models (e.g., GPT~\cite{brown2020language} and BERT~\cite{devlin2018bert}) have achieved unparalleled performance on various natural language processing (NLP) tasks.

Whereas in the EDA domain, despite all the recent efforts in learning-based solutions~\cite{huang2021machine}, obtaining a \textit{general and effective circuit representation} that serves as the basis for solving various EDA tasks has not been addressed yet. In this work, we take the first step towards this direction by introducing a novel GNN-based solution for the representation learning of logic gates, namely \textit{DeepGate}, which is aware of the logic computation procedure and the structural information of combinational circuits.

Naturally, logic circuits can be modeled as directed acyclic graphs (DAGs), in which logic gates appear in a specific topological order. Therefore, one could collect many logic circuits and resort to existing DAG-GNN architecture~\cite{zhang2019dvae,thost2021directed} to learn the node embedding for each logic gate with some supervision tasks (e.g., Boolean satisfiability~\cite{amizadeh2018learning}). However, we argue that such straightforward solutions cannot effectively extract information from circuit graphs.

Firstly, logic circuits could follow different design styles and use diverse technology libraries containing various logic gate types, leading to heterogeneous circuit graphs with mixed distributions that are challenging to learn. In DeepGate, we propose to conduct learning on a general intermediate representation of logic circuits, i.e., \textit{and-inverter graph} (AIG), with the help of logic synthesis tools~\cite{brayton2010abc}. The benefits are twofold: (i). Such a unified format constrains the circuit graph distribution without changing circuit functionalities. All the transformed circuits only feature two types of logic gates (i.e., 2-input AND gate and inverter); (ii). The logic synthesis procedure naturally introduces a strong inductive bias of practical circuits for effective learning with GNN models.


Secondly, the effectiveness of representation learning heavily relies on supervision tasks. For example, when pre-training CV models, the image class labels of the ImageNet dataset serve as the cornerstone. In contrast, considering the difficulty in annotating a textual dataset as large as ImageNet, self-supervised tasks are used instead in NLP pre-trained model development. For effective circuit representation learning, we propose to use the signal probability (i.e., the probability of being logic ‘1’) for every node as rich supervision because it embeds the genuine logic relationship of each node in the circuits. To be specific, we perform logic simulations with a large amount of random patterns to obtain faithful probability values for supervision.

Last but most important, existing GNN models are general solutions designed to extract information from all kinds of graphs, while circuit graphs are a unique type of graph with logic relationships between nodes. In this work, we design a dedicated GNN model for circuit graphs, significantly enhancing the learning effectiveness. 

We summarize the contributions of this work as follows:

\begin{itemize}
    \item To the best of our knowledge, DeepGate is the first work for the \textit{general and effective circuit representation learning} problem. Specially, we propose a novel design flow to tackle this problem: (i). circuit transformation in AIG form; (ii). supervision with logic-simulated probabilities; (iii). representation learning with a dedicated GNN model for circuit graphs. 
    \vspace{5pt}
    \item We propose a novel GNN model for circuit graphs that exploits unique circuit properties, including \textit{attention mechanisms} that mimic the logic computation procedure and \textit{reversed propagation layers} that consider logic implication effects.
    \vspace{5pt}
    \item Reconvergence structures are inevitable due to logic sharing in multi-level logic networks, and they are the main challenges for logic analysis~\cite{roberts1987algorithm}. We treat them as first-class citizen and introduce novel solutions in our GNN model. 
\end{itemize}

We learn the representations of logic gates with many small subcircuits extracted from benchmark circuits. Experimental results performed on large circuits show the efficacy and generalization capability of DeepGate. We organize the remainder of this paper as follows. We review related works in Section~\ref{sec:related}. Section~\ref{sec:method} introduces the DeepGate architecture, while in Section~\ref{sec:experiment}, we present the experimental results on various circuits. Finally, Section~\ref{sec:conclusion} concludes this paper.

\section{Related Works}
\label{sec:related}
\vspace{5pt}

\begin{figure}[t!]
	\centering
	\includegraphics[width=0.8\linewidth]{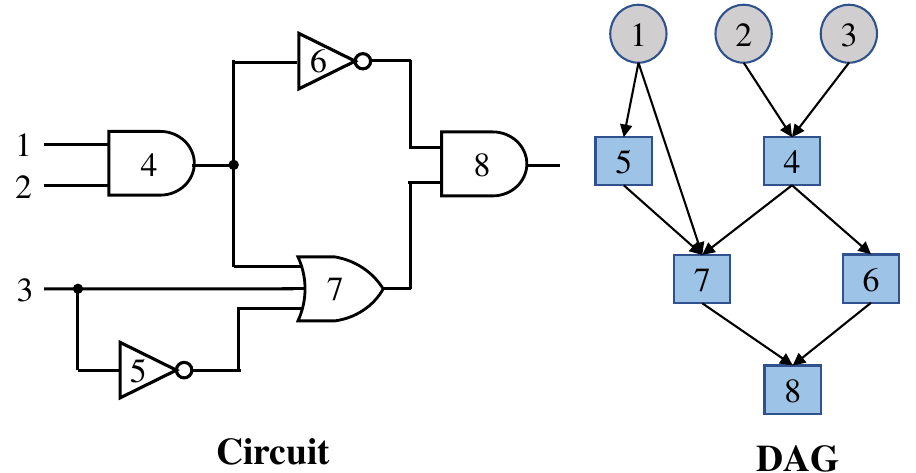}
	\caption{The Circuit Representation as DAG}
	\label{FIG:DAG}
\end{figure}

\subsection{Graph Neural Networks}
Graph neural networks~\cite{hamilton2017inductive,kipf2016semi} have received a lot of attention for their effectiveness in modeling non-structured data. By learning vectorial representations on graphs via feature propagation and aggregation, GNNs show convincing results in various domains~\cite{hu2020open,gilmer2017neural, wu2018socialgcn}. The most popular GNN model employs a message-passing neural network architecture, which computes representation/hidden states $\mathbf{h}_v^\ell$ for node $v$ in a graph $\mathcal{G}$ at every layer $\ell$ and a final graph representation $\mathbf{h}_\mathcal{G}$, as in \cite{gilmer2017neural}: 
\begin{equation}
\hspace{-1mm}
    \footnotesize
    \mathbf{h}_v^\ell = \text{COMBINE}^\ell(\mathbf{h}_v^{\ell-1}, \text{AGGREGATE}^\ell(\{\mathbf{h}_u^{\ell-1} | u\in \mathcal{N}(v)\})), \ell = 1,.,L
    \label{eq:gnn-edge}
\end{equation}
\begin{equation}
    \small
    \mathbf{h}_\mathcal{G} = \text{READOUT}(\{\mathbf{h}_v^L, v\in \mathcal{V}\})
    \label{eq:gnn-graph}
\end{equation}
wherein $N(v)$ denotes neighboring nodes of node $v$ and $L$ is the number of layers. The parameterized function $\text{AGGREGATE}^\ell$  aggregates messages from neighboring nodes $N(v)$, and $\text{COMBINE}^\ell$ obtains an updated hidden state after aggregation. Finally, the function $\text{READOUT}^\ell$   retrieves the states of all nodes $\mathcal{V}$  and produces the graph neural representation. A notable GNN architecture is the graph attention network (GAT) \cite{velikovi2017graph} that considers the importance of different neighbors during aggregation.

Directed acyclic graphs (DAGs) are a special type of graphs, yet broadly seen across many domains, including circuit modeling (see Fig.~\ref{FIG:DAG}). Recently, few studies have been dedicated to DAG-GNN designs~\cite{zhang2019dvae,thost2021directed}, which propagate the message following the topological ordering between nodes and only consider the predecessors in the $\text{AGGREGATE}^\ell$ function, as demonstrated in Equation~\eqref{eq:dag-edge}.
\begin{equation}
\hspace{-0.5mm}
    \footnotesize
    \mathbf{h}_v^\ell = \text{COMBINE}^\ell(\mathbf{h}_v^{\ell-1}, \text{AGGREGATE}^\ell(\{\mathbf{h}_u^{\ell} | u\in \mathcal{P}(v)\})), \ell = 1,...,L
    \label{eq:dag-edge}
\end{equation}
The major difference between Eq.~\eqref{eq:dag-edge} and Eq.~\eqref{eq:gnn-edge} is that in DAG-GNN, the aggregation function for $v$ will be only executed after all of its predecessors' hidden states have already been computed.
 
 Besides stacking $L$ layers to increase the depth of the network, one can also apply the same model for $T$ times in the \textit{recurrent} fashion to generate the final embedding~\cite{ amizadeh2018learning}:
\begin{equation}
    \footnotesize
    \mathbf{h}_v^t = \text{COMBINE}(\mathbf{h}_v^{t-1}, \text{AGGREGATE}(\{\mathbf{h}_u^{t} | u\in \mathcal{P}(v)\})), t = 1,...,T
    \label{eq:recdag-edge}
\end{equation}

Using the taxonomy defined in~\cite{wu2020comprehensive}, we name the two variants of DAG-GNNs described in Equations~\eqref{eq:dag-edge}--\eqref{eq:recdag-edge} as \textit{DAG-ConvGNNs} and \textit{DAG-RecGNNs}, respectively.

\subsection{GNN-Based Solutions for EDA Problems}
Existing GNN-based EDA solutions use an end-to-end flow for specific EDA tasks wherein the labels are usually extracted from commercial EDA tools. 

To the best of our knowledge, the first GNN-based EDA technique is applied to the test point insertion (TPI) problem, which is formulated as a node binary classification problem and solved with a graph convolutional network~\cite{ma2019high}. 
The ground-truth labels are collected from commercial TPI tools, revealing whether a particular node is "easy to observe" or not. CongestionNet~\cite{kirby2019congestionnet} models the circuit as an undirected graph and trains a GAT model to predict the congestion of the final physical design on a per-cell basis. 
GRANNITE~\cite{zhang2020grannite} conducts power estimation using a DAG-GNN model. Gate netlists are mapped onto graphs with per-node (gate) and per-edge (net) features. They achieved good accuracy (less than 5.5\% error across a diverse set of benchmarks) for fast (<1 second) average power estimation on designs up to 50k gate. 
Recently, \cite{xie2021net} proposes a GAT-based model named \(Net^2\) for pre-placement net length estimation. 
To solve a particular EDA problem, the above techniques typically pre-compute many node/edge features (e.g., SCOAP testability measures in \cite{ma2019high}) 
and use existing GNN models to aggregate these features for solution findings. Consequently, the learned node features cannot be transferred among related tasks, despite using the same circuit graphs as inputs. More importantly, an effective representation for circuits should be aware of their logic functions. However, existing solutions ignore it and only consider the structural information in their learning procedure. 

Motivated by the above, we propose to learn a general and effective circuit representation without pre-computing any specific features, as detailed in the following section.

\vspace{10pt}
\section{Proposed Solution}
\label{sec:method}


\begin{figure*}[t!]
	\centering
	\includegraphics[width=\linewidth]{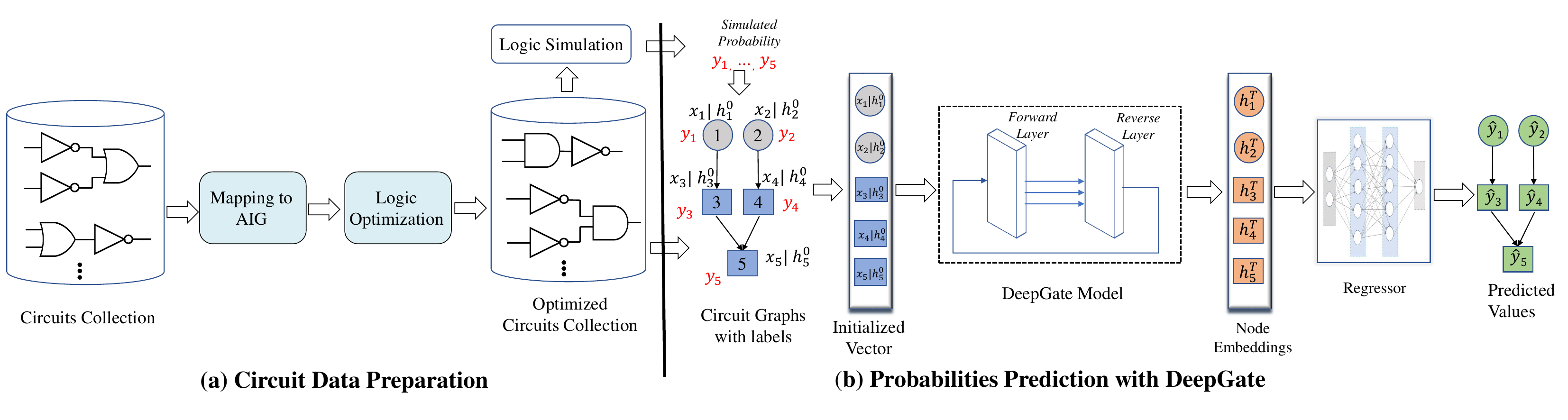}
	\caption{The Overview of DeepGate.} 
	\label{FIG:Arch}
\end{figure*}

\subsection{Overview of DeepGate}

Figure~\ref{FIG:Arch} presents the overview of the proposed DeepGate solution, consisting of two stages for the neural representation learning of logic gates: 

\begin{itemize}
    \item \textbf{Circuit Data Preparation:} Given a pool of circuit designs, we use logic synthesis tools to transform them into a unified AIG format. We then perform logic simulations on the circuits with sufficient random patterns to obtain the signal probability (i.e., the probability of node being logic `1`) on every node as supervision. We elaborate the details in Section~\ref{sec:data-pre}.
    \vspace{5pt}
    \item \textbf{Probabilities Prediction with DeepGate:} Given a circuit dataset and the logic-simulated probabilities as the supervision task, we introduce a novel GNN model dedicated for circuit graph analysis to learn the neural representations of logic gates, as detailed in Section~\ref{sec:model}.
\end{itemize}

\subsection{Circuit Data Preparation}\label{sec:data-pre}
Some circuits are at the register-transfer level, while others are gate-level netlists mapped with various libraries. Such heterogeneity across circuits is a challenge for GNN model development. To tackle this problem, we resort to the logic synthesis tool ABC~\cite{brayton2010abc} and transform all circuits into the unified AIG format. If the original circuit is too large, we extract small sub-circuits with circuit sizes ranging from $30$ to $3k$ gates. Note that, we test the effectiveness of DeepGate on much larger circuits for its generalization capabilities.

The benefits of such circuit pre-processing flow include: (i).~Only two logic gate types (i.e., 2-input \textit{AND} gate and 1-input \textit{Not} gate) are considered, which would dramatically reduce representation learning difficulty; (ii).~Applying logic synthesis introduces strong relational inductive bias into the resulting circuit graphs; (iii).~The constraint on circuit size facilitates efficient GNN training with both reduced sizes of circuit graphs and less time for preparing supervision labels. 

There are many possibilities to annotate a circuit, e.g., the satisfiability of the circuit~\cite{amizadeh2018learning}. However, a good supervision task should satisfy the following condition: the labels should be easily obtained while retaining rich information for both the logic function and the structural information of the circuits. In DeepGate, we propose to use the signal probability on every node as supervision, which satisfies the above requirements: (i). It is relatively easy to obtain highly-accurate probability values by running logic simulations on many random input patterns, especially when the circuit size is limited; (ii). A unique yet important property of logic circuits that makes circuit analysis challenging is the reconvergence structures, and logic simulation is arguably the only way to obtain the actual value for such structures; (iii). The logic probability of each gate itself plays an essential role in many EDA tasks. 

\subsection{GNN Model in DeepGate}\label{sec:model}

Given circuit graphs in AIG form, the objective of our GNN model is to estimate the probability of every node such that it would be as close to the genuine signal probability as possible. 
Different from existing DAG-ConvGNNs~\cite{thost2021directed,zhang2019dvae} and DAG-RecGNNs~\cite{amizadeh2018learning} models that focus on learning the topological information in the graph, 
DeepGate is designed to learn 
both the circuit structural information and the computational behaviour of logic circuits, and embed them as vectors on every logic gate.

We now elaborate on the detailed GNN model design in DeepGate. Given a circuit graph $\mathcal{G}$, we embed the gate type of each node $v$ with one-hot encoding in $\mathbf{x}_v$. To be specific, as only primary inputs (PIs), \textit{And} gates and \textit{Not} gates are present in AIGs, we assign a $3$-d vector for each node according to its gate type. It should be noted that instead of relying on the probability-based measurements in previous works~\cite{zhang2020grannite,ma2019high}, our model only requires gate type information for the representation learning. We also have hidden states $\mathbf{h}_v$ 
for every node, which is initialized randomly. Given these, DeepGate resorts to attention-based aggregation design~\cite{thost2021directed,velikovi2017graph} and the gated recurrent unit (GRU)~\cite{zhang2019dvae} as the update function.

\vspace{5pt}
\textit{Aggregation.} We use the attention mechanism in the additive form to instantiate the AGGREGATE function in Equation~\eqref{eq:dag-edge}, wherein the aggregated message $\mathbf{m}_v^t$ for a node $v$ at $t^{th}$ iteration is computed by:
\begin{equation}
\small
    \mathbf{m}_v^t = \sum_{u \in \mathcal{P}(v)}  \alpha_{uv}^t  \mathbf{h}_u^t
\text{\quad where\quad} \alpha_{uv}^t = \mathop{softmax}\limits_{u \in \mathcal{P}(v)} (w_1^\top \mathbf{h}_v^{t-1}+ w_2^\top \mathbf{h}_u^t)
\label{eq:attn}
\end{equation}
where $\alpha_{uv}^t$ is a weighting coefficient that is computed by following the query-key design as in usual attention mechanisms. To be specific, $\mathbf{h}_v^{t-1}$ serves as \textit{query}, and representation of predecessors from current iteration $t$, $\mathbf{h}_u^t$, serves as \textit{key}. The intuition behind using the attention mechanism for aggregation is that when we do the logic computation in digital circuits, the controlling value of a logic gate determines the output of that gate. Therefore, controlling values are far more important than non-controlling values. To mimic this behaviour, the attention mechanism can learn to assign high weights for controlling inputs of gates and give less importance to the rest of the inputs.

\vspace{5pt}
\textit{Combine.} We then use the GRU to instantiate the COMBINE function for updating the hidden state of target node $v$:
\begin{equation}
    \mathbf{h}_v^t = GRU([\mathbf{m}_v^t,\mathbf{x}_v], \mathbf{h}_v^{t-1})
\end{equation}
wherein $\mathbf{m}_v^t$, $\mathbf{x}_v$ are concatenated together and treated as input, while $\mathbf{h}_v^{t-1}$ is the past state of GRU. 

On the one hand, DeepGate adopts the recursive DAG-GNNs functional defined in Equation~\eqref{eq:recdag-edge}.
The reasons for using the recurrent architecture are two-fold: (i). 
it is unrealistic for GNNs to capture the circuit's functional and structural information with a single forward propagation; (ii). the recurrent learning procedure facilitates reaching stabilized node embeddings quickly. 

On the other hand, our proposed GNN model differs from previous DAG-GNNs~\cite{amizadeh2018learning,thost2021directed,zhang2019dvae} that initialize $\mathbf{h}_v^0$ as $\mathbf{x}_v$ and treat the aggregated message as the state of recurrent function. In contrast, we fix the gate type information of nodes $\mathbf{x}_v$ as inputs for all iterations. Such employment can avoid the information vanishing of gate properties during the long-term recursive propagation.

\vspace{5pt}
\textit{Reversed Propagation Layer.}
In DeepGate, we also consider backward information propagation, i.e., processing the graph in reversed topological order. One of the main reasons to introduce the backward layers in our framework is that logic implication and backtracking in the reversed order can be highly useful for predicting the states of nodes. It also helps stabilize training, as proved in sequence-to-sequence learning tasks~\cite{sutskever2014sequence}.

\vspace{5pt}
\textit{Regressor.} After $T$ iterations, we pass the hidden states of nodes $\mathbf{h}_v^T$ into a multi-layer perceptron (MLP), which computes a single scalar for every node to regress the simulated probabilities. The weights of MLP are shared for nodes with the same gate types. We train the network to minimize the \textit{L1} loss between the prediction $\hat{y}_v$ and the true probability $y_v$.

\begin{figure}[t!]
 	\centering
 	\includegraphics[width=0.8\linewidth]{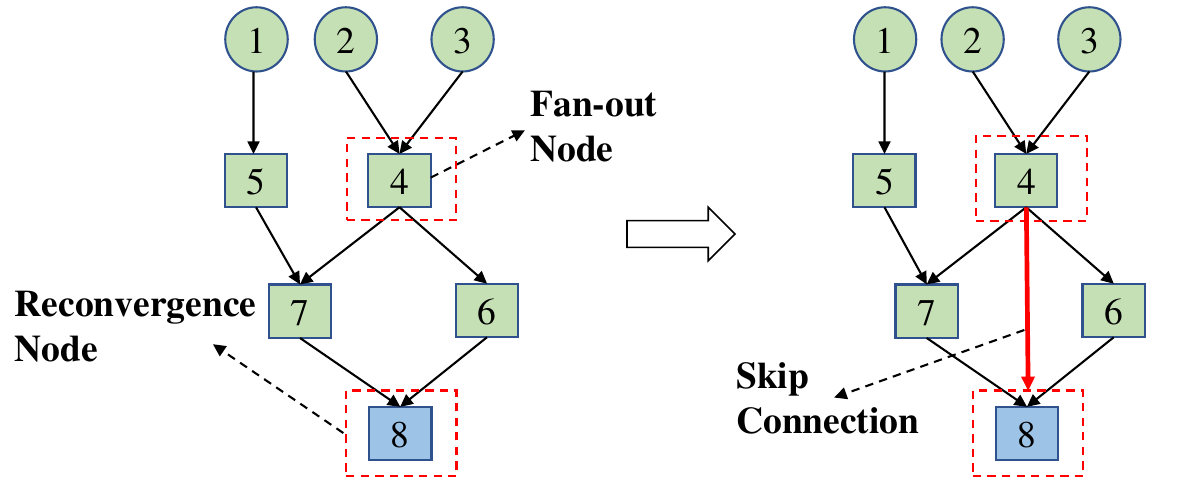}
	\caption{Information Propagation at Reconvergence Node using Skip Connection}	\label{fig:SC}
	\vspace{-5pt}
\end{figure}

\subsection{Skip Connection for Reconvergence Structure with Positional Encoding}
In the previous section we have described the core components of DeepGate necessary for predicting the logic probabilities of nodes. However, the logic inference on reconvergence nodes is different from normal nodes and such structures are inevitable due to logic sharing in multi-level logic networks. Hence, they are the main challenge for logic probability analysis. 
To accommodate their impact, we introduce the improvement into DeepGate to enable special processing for reconvergence nodes as shown in Figure~\ref{fig:SC}. 

Firstly, we maintain the information of reconvergence nodes during circuit data preparation, including its corresponding source fan-out node, and the logic level difference between the source nodes and reconvergence nodes. Secondly, we add direct edges between the fan-out node and the reconvergence node, named \textit{skip connection} here. The new edges facilitate the information exchange from fan-out nodes to reconvergence nodes. Last but not least, we leverage the positional encoding technique~\cite{vaswani2017attention} to differentiate the skip connection and the normal connection. To be specific, the function $\gamma(D)$ is a mapping of logic level difference $D$ between source fan-out node and reconvergence node into a higher dimensional space $\mathbb{R}^{2L}$:
\begin{equation}
    \footnotesize
    \label{eq:positional}
    \gamma(D) = (\sin(2^0\pi D), \cos(2^0\pi D), \cdots, \sin(2^{L-1}\pi D), \cos(2^{L-1}\pi D))
\end{equation}
The impact of the fanout node on the reconvergence nodes depends upon the distance between them. Generally speaking, the longer the distance is, the lesser impact it has on the reconvergence node. The above function can induce the knowledge of how much fanout node can impact the result of reconvergence node into the model. We assign the encoded vector as the edge attributes to skip connection and incorporate it into the coefficient calculation described in Equation~\eqref{eq:attn} as the third input. 




\vspace{10pt}
\section{Experiments}
\label{sec:experiment}
\vspace{10pt}

\subsection{Datasets}\label{sec:dataset}
\vspace{5pt}

\begin{table}
\centering
\caption{The Statistics of Circuit Training Dataset}
\label{Tab:TrainSet}

\begin{tabular}{llll} 
\toprule
Benchmark & \#Subcircuits & \#Node & \#Level\\ 
\midrule
EPFL      & 828          & [52--341]      & [4--17] \\
ITC99     & 7,560          & [36--1,947]    & [3--23]   \\
IWLS      & 1,281          & [41--2,268]   & [5--24]   \\
Opencores & 1,155          & [51--3,214]   & [4--18]  \\
\midrule
\textbf{Total}   & \textbf{10,824} & \textbf{[36--3,214]} & \textbf{[3--24]} \\
\bottomrule
\end{tabular}
\end{table}
We extract many sub-circuits from four circuit benchmark suites: ITC’99~\cite{ITC99}, IWLS’05~\cite{albrecht2005iwls}, EPFL~\cite{EPFLBenchmarks} and OpenCore~\cite{takeda2008opencore}, and follow the circuit data preparation flow described in Section~\ref{sec:data-pre} to transform all circuits into a unified AIG format. We conduct logic simulations with up to $100k$ random input patterns to obtain an accurate signal probability on every node.

Table~\ref{Tab:TrainSet} presents the statistics of the circuit dataset.
\#Subcircuits shows the total number of subcircuits extracted from each benchmark. As shown in the table, the constructed circuit dataset covers circuit sizes ranging from tens to thousands of nodes with different logic levels. Finally, there are 
$10,824$ circuits in total, and we create $90/10$ training/test splits for model training and evaluation.

\subsection{Evaluation Metric and Baselines}
To evaluate the performance of different GNN models, we calculate the average value of the absolute differences between the simulated probability $y_v$ and the predicted $\hat{y}_v$ from DeepGate for all nodes $\mathcal{V}$ in the circuits, as shown in equation~\eqref{eq:loss}. The smaller the value is, the better the model performs.
\vspace{5pt}
\begin{equation}
    \label{eq:loss}
    Avg.~Prediction~Error = \frac{1}{N} \sum_{v\in\mathcal{V}}{\left| y_v - \hat{y}_v \right|}
\end{equation}

We consider three GNN models: GCN, DAG-ConvGNN, and DAG-RecGNN. \textit{GCN} model treats the circuit graphs as undirected graphs in representation learning. DAG-ConvGNN model follows the settings defined in Equation~\eqref{eq:dag-edge}. For DAG-RecGNN model, we adopt the same COMBINE function and the reversed propagation layer design in DeepGate, as depicted in Section~\ref{sec:model}. As for the GNN model in DeepGate, it contains additional attention mechanism and skip connection (SC).  Under every setting, we evaluate 4 different aggregator designs, which include representative works for DAG learning, i.e., Convolutional Sum (abbreviated as Conv. Sum)~\cite{selsam2018learning}, Attention~\cite{velikovi2017graph, thost2021directed}, GatedSum~\cite{zhang2019dvae} and DeepSet~\cite{amizadeh2018learning}. 

In order to make the comparison fair, we instantiate all models with $d=64$ for the node hidden states $\mathbf{h}_v$ and design the other parameterized functions to have a similar number of tunable parameters. For DAG-RecGNNs and our DeepGate model, a forward layer is followed by a reversed layer, and $T=10$ iterations of message passing are performed to obtain the final embeddings. We choose $L=8$ in Equation~\eqref{eq:positional} for the skip connection setting. For training, all models are optimized for $60$ epochs using the ADAM optimizer with a learning rate of $1\times10^{-4}$. We use the topological batching technique introduced in~\cite{thost2021directed} to accelerate the training.

\begin{table}
\centering
\caption{The Performance Comparison of DeepGate with other GNN models for Logic Probability Prediction}
\label{Tab:Comparison}
\begin{tabular}{lcc} 
\hline
Model             & Aggregator                                       & Avg. Prediction Error  \\ 
\hline
GCN               & Conv. Sum                                        & 0.1386                 \\
                  & Attention                                        & 0.1840                 \\
                  & DeepSet                                          & 0.2541                 \\
                  & GatedSum                                         & 0.1995                 \\ 
\hline
DAG-ConvGNN       & Conv. Sum                                        & 0.2215                 \\
                  & Attention                                        & 0.2398                 \\
                  & DeepSet                                          & 0.2431                 \\
                  & GatedSum                                         & 0.2333                 \\ 
\hline
DAG-RecGNN        & Conv. Sum                                        & 0.0328                 \\
(T=10)            & DeepSet                                          & 0.0302                 \\
                  & GatedSum                                         & 0.0329                 \\ 
\hline
\textbf{DeepGate} & Attention w/o SC                                 & \textbf{0.0234 }       \\
(T=10)            & Attention w/ SC & \textbf{0.0204 }       \\
\hline
\end{tabular}
\end{table}

\subsection{Probability Prediction}~\label{sec:prob}
\subsubsection{Comparison of DeepGate with Baseline Solutions}

Table~\ref{Tab:Comparison} compares DeepGate with other baseline solutions in terms of prediction error. From this table, we have several observations: First, both GCN and DAG-ConvGNN are subject to poor performance for probability prediction, mainly due to their lack of ability to model the computational behaviours of circuits. For instance, the best GCN model, equipped with Conv. Sum, gives $0.1386$ of prediction error, which in turn is even higher than the worst performing model of DAG-RecGNN. Therefore, only by incorporating the logical ordering into the model design and conducting the propagation recurrently, will make the model perform well. It shows the advantage of DAG-RecGNN implementation with dedicated recurrent scheme and reversed layer design discussed in Section~\ref{sec:model} over simpler GNN architectures. 
Second, among all models, DeepGate with attention alone achieves significant prediction error reduction. It brings $22.76\%$ relative improvement compared with the best baseline solution, which is the DAG-RecGNN model equipped with DeepSet aggregator. Hence, using the dedicated attention mechanism benefits logic representation learning. 
Third, equipped with skip connection design, DeepGate can further reduce the prediction error from $0.0234$ to $0.0204$, which reveals the efficacy of introducing the reconvergence knowledge into the model design. To summarize, with only the gate type information and the connectivity between gates, DeepGate learns to predict highly-accurate probabilities for logic gates.

As we observe that DAG-RecRNN with DeepSet aggregator (abbreviated as \textit{DeepSet} for the following discussion) performs better than the other baselines, in later results, we only compare DeepGate (w/ skip connection) with it.


\subsubsection{Results on Large Circuits}

Furthermore, we evaluate DeepGate on five circuit designs that are substantially larger than the circuits it saw during training. The circuit statistics and the prediction error of both DeepGate and DeepSet are shown in Table~\ref{Tab:large-comparison}. The number of gates in these designs is two orders of magnitude more than that of the training circuits. 

We can observe that DeepGate achieves similar prediction accuracy as that on small circuits, and it outperforms DeepSet in these large circuits by a large margin. Such results clearly demonstrate the generalization capability of DeepGate.  In particular, DeepGate achieves $73.56\%$ prediction error reduction on \textit{Arbiter}. 
This is because, the Arbiter circuit is designed to accommodate access from multiple requests, and it contains repetitive logic units with many reconvergence structures. As DeepGate treats such structures as a first-class citizen in the GNN model, it can generate much more accurate predictions. 




\begin{table}
\centering
\renewcommand\tabcolsep{2.0pt}
\caption{The Performance Comparison of DeepGate and DeepSet on Five Large Circuits}
\label{Tab:large-comparison}
\begin{tabular}{@{}llllll@{}}
\toprule
Design  & \#Nodes & Levels & DeepSet & DeepGate & Reduction \\ \midrule
Arbiter      & 23.7K   & 173    & 0.0277  & 0.0073   & 73.56\%   \\
Squarer      & 36.0K   & 373    & 0.0495  & 0.0346   & 30.16\%   \\
Multiplier      & 47.3K   & 521    & 0.0220  & 0.0159   & 27.94\%   \\
80386 Processor      & 13.2K   & 122    & 0.0534  & 0.0387   & 27.56\%   \\
Viper Processor      & 40.5K   & 133    & 0.0520  & 0.0389   & 25.18\%   \\  \bottomrule
\end{tabular}
\end{table}

\vspace{5pt}
\subsection{Discussion}
\vspace{10pt}
\subsubsection{Effectiveness of Circuit Transformation}

DeepGate uses the logic synthesis tool to transform the circuits from different sources into unified AIG forms. One may wonder the performance of DeepGate if the network is directly trained on the original circuits, wherein other gate types (e.g., XOR, NAND, NOR, and OR) are also included. To investigate the effectiveness of the circuit transformation in DeepGate, we conduct the controlled experiments on EPFL and IWLS benchmarks, as shown in Table~\ref{tab:epfl}. 

Take EPFL as an example, we extract $375$ sub-circuits from the original designs, and develop two versions: the ones with the original $6$ gates types and the other with AIG transformation. For each version, we train the DeepGate model from scratch. The only difference is that for the former version of the dataset, we assign $7$-d one-hot encoding for the node feature $\mathbf{x}_v$. As can be observed from Table~\ref{tab:epfl}, DeepGate trained on AIGs performs better than the one trained on the original circuits by a large margin ($33.94\%$ relative prediction error reduction on EPFL). The same observation can be obtained from the results on IWLS circuits. 

Such improvements originate from the benefit of circuit transformation because when only two logic gate types are considered, representation learning difficulty is reduced dramatically without any impact on circuit functionalities. Also, we manually check the usage frequency of different gate types in the original formats, and observe that some gates types (e.g., XOR and NAND) are used much less frequently. Such in-balanced gate distributions may lead to insufficient training, causing higher prediction errors. 

Additionally, we directly apply the pre-trained DeepGate model on the merged AIG dataset, as shown in Section~\ref{sec:dataset} for comparison. As can be observed, DeepGate trained on the dataset consisting of different benchmarks can further reduce the prediction errors by $51.37\%$. It supports the claim that unifying different circuit designs into a common intermediate representation can help the model learn a better representation of logic gates.

\begin{table}[t!]
\centering
\caption{The Performance of DeepGate with and without Circuit Transformation} \label{tab:epfl}
\begin{tabular}{@{}lccc@{}}
\toprule
  & w/o Tran. & w/ Tran.    & Pre-trained \\ \midrule
 EPFL & 0.0442     & 0.0292 & 0.0142        \\ 
 IWLS & 0.0447     & 0.0342 & 0.0209        \\
 \bottomrule
\end{tabular}
\end{table}

\vspace{10pt}
\subsubsection{Impact of Recurrence Iterations}

During the inference phase, the number of iterations $T$ can be set as different values. The higher the value, the higher the computational cost. 

We enumerate different values for $T$, ranging from $1$ to $50$. We observe that our GNN model is able to decrease the prediction loss as $T$ increases. However, the prediction error converges quickly at around $T=10$, despite the circuit size. Such experimental results further demonstrate the scalability of the proposed DeepGate solution. 



\vspace{5pt}
\section{Conclusion and Future Work}
\label{sec:conclusion}
\vspace{5pt}

This paper proposes DeepGate, a novel representation learning solution that effectively embeds both logic functions and structural information of a circuit as vectors on each gate. In DeepGate, we construct easy-to-learn circuit graphs by transforming circuits into unified AIG format and introduce a novel GNN model with circuit knowledge as priors for effective representation learning. Using informative signal probability as supervision tasks on small sub-circuits, we show DeepGate can generalize to large circuits with accurate predictions without any pre-computed features.

While showing promising results, the current DeepGate model is still in its infancy. For example, we could introduce other informative supervision tasks (e.g., logic inference and Boolean satisfiability) to achieve better representations for logic gates. We could also add more circuits for training to build a large-scale foundation model for logic circuits~\cite{bommasani2021opportunities}. Moreover, in our future work, we plan to apply the representations learned in DeepGate onto many downstream EDA tasks (e.g., power estimation, logic reduction, and equivalence checking). These tasks are directly related to signal probability analysis, and we believe DeepGate can achieve satisfactory results without much effort in finetuning the model.
\balance
\bibliographystyle{IEEEtran}
\bibliography{reference}

\begin{thebibliography}{10}
\providecommand{\url}[1]{#1}
\csname url@samestyle\endcsname
\providecommand{\newblock}{\relax}
\providecommand{\bibinfo}[2]{#2}
\providecommand{\BIBentrySTDinterwordspacing}{\spaceskip=0pt\relax}
\providecommand{\BIBentryALTinterwordstretchfactor}{4}
\providecommand{\BIBentryALTinterwordspacing}{\spaceskip=\fontdimen2\font plus
\BIBentryALTinterwordstretchfactor\fontdimen3\font minus
  \fontdimen4\font\relax}
\providecommand{\BIBforeignlanguage}[2]{{%
\expandafter\ifx\csname l@#1\endcsname\relax
\typeout{** WARNING: IEEEtran.bst: No hyphenation pattern has been}%
\typeout{** loaded for the language `#1'. Using the pattern for}%
\typeout{** the default language instead.}%
\else
\language=\csname l@#1\endcsname
\fi
#2}}
\providecommand{\BIBdecl}{\relax}
\BIBdecl

\bibitem{huang2021machine}
G.~Huang, J.~Hu, Y.~He, J.~Liu, M.~Ma, Z.~Shen, J.~Wu, Y.~Xu, H.~Zhang,
  K.~Zhong \emph{et~al.}, ``Machine learning for electronic design automation:
  A survey,'' \emph{TODAES}, vol.~26, no.~5, pp. 1--46, 2021.

\bibitem{kipf2016semi}
T.~N. Kipf and M.~Welling, ``Semi-supervised classification with graph
  convolutional networks,'' \emph{arXiv preprint arXiv:1609.02907}, 2016.

\bibitem{hamilton2017inductive}
W.~L. Hamilton, R.~Ying, and J.~Leskovec, ``Inductive representation learning
  on large graphs,'' in \emph{NIPS}, 2017, pp. 1025--1035.

\bibitem{kirby2019congestionnet}
R.~Kirby, S.~Godil, R.~Roy, and B.~Catanzaro, ``Congestionnet: Routing
  congestion prediction using deep graph neural networks,'' in \emph{2019
  VLSI-SoC}.\hskip 1em plus 0.5em minus 0.4em\relax IEEE, pp. 217--222.

\bibitem{ma2019high}
Y.~Ma, H.~Ren, B.~Khailany, H.~Sikka, L.~Luo, K.~Natarajan, and B.~Yu, ``High
  performance graph convolutional networks with applications in testability
  analysis,'' in \emph{DAC}, 2019, pp. 1--6.

\bibitem{han2021pre}
X.~Han, Z.~Zhang, N.~Ding, Y.~Gu, X.~Liu, Y.~Huo, J.~Qiu, L.~Zhang, W.~Han,
  M.~Huang \emph{et~al.}, ``Pre-trained models: Past, present and future,''
  \emph{AI Open}, 2021.

\bibitem{krizhevsky2012imagenet}
A.~Krizhevsky, I.~Sutskever, and G.~E. Hinton, ``Imagenet classification with
  deep convolutional neural networks,'' \emph{NIPS}, pp. 1097--1105, 2012.

\bibitem{minaee2021image}
S.~Minaee, Y.~Y. Boykov, F.~Porikli, A.~J. Plaza, N.~Kehtarnavaz, and
  D.~Terzopoulos, ``Image segmentation using deep learning: A survey,''
  \emph{IEEE Transactions on Pattern Analysis and Machine Intelligence}, 2021.

\bibitem{ren2015faster}
S.~Ren, K.~He, R.~Girshick, and J.~Sun, ``Faster r-cnn: Towards real-time
  object detection with region proposal networks,'' \emph{NIPS}, pp. 91--99,
  2015.

\bibitem{brown2020language}
T.~B. Brown, B.~Mann, N.~Ryder, M.~Subbiah, J.~Kaplan, P.~Dhariwal,
  A.~Neelakantan, P.~Shyam, G.~Sastry, A.~Askell \emph{et~al.}, ``Language
  models are few-shot learners,'' \emph{arXiv preprint arXiv:2005.14165}, 2020.

\bibitem{devlin2018bert}
J.~Devlin, M.-W. Chang, K.~Lee, and K.~Toutanova, ``Bert: Pre-training of deep
  bidirectional transformers for language understanding,'' \emph{arXiv preprint
  arXiv:1810.04805}, 2018.

\bibitem{zhang2019dvae}
M.~Zhang, S.~Jiang, Z.~Cui, R.~Garnett, and Y.~Chen, ``D-vae: A variational
  autoencoder for directed acyclic graphs,'' 2019.

\bibitem{thost2021directed}
V.~Thost and J.~Chen, ``Directed acyclic graph neural networks,'' in
  \emph{ICLR}, 2021.

\bibitem{amizadeh2018learning}
S.~Amizadeh, S.~Matusevych, and M.~Weimer, ``Learning to solve circuit-{SAT}:
  An unsupervised differentiable approach,'' in \emph{ICLR}, 2019.

\bibitem{brayton2010abc}
R.~Brayton and A.~Mishchenko, ``Abc: An academic industrial-strength
  verification tool,'' in \emph{CAV}.\hskip 1em plus 0.5em minus 0.4em\relax
  Springer, 2010, pp. 24--40.

\bibitem{roberts1987algorithm}
M.~Roberts and P.~Lala, ``Algorithm to detect reconvergent fanouts in logic
  circuits,'' \emph{IEEE Proceedings Computers and Digital Techniques}, 1987.

\bibitem{hu2020open}
W.~Hu, M.~Fey, M.~Zitnik, Y.~Dong, H.~Ren, B.~Liu, M.~Catasta, and J.~Leskovec,
  ``Open graph benchmark: Datasets for machine learning on graphs,''
  \emph{arXiv preprint arXiv:2005.00687}, 2020.

\bibitem{gilmer2017neural}
J.~Gilmer, S.~S. Schoenholz, P.~F. Riley, O.~Vinyals, and G.~E. Dahl, ``Neural
  message passing for quantum chemistry,'' in \emph{ICML}, 2017, pp.
  1263--1272.

\bibitem{wu2018socialgcn}
L.~Wu, P.~Sun, R.~Hong, Y.~Fu, X.~Wang, and M.~Wang, ``Socialgcn: An efficient
  graph convolutional network based model for social recommendation,''
  \emph{arXiv preprint arXiv:1811.02815}, 2018.

\bibitem{velikovi2017graph}
P.~Veličković, G.~Cucurull, A.~Casanova, A.~Romero, P.~Liò, and Y.~Bengio,
  ``Graph attention networks,'' \emph{ICLR}, 2017.

\bibitem{wu2020comprehensive}
Z.~Wu, S.~Pan, F.~Chen, G.~Long, C.~Zhang, and S.~Y. Philip, ``A comprehensive
  survey on graph neural networks,'' \emph{IEEE transactions on neural networks
  and learning systems}, vol.~32, no.~1, pp. 4--24, 2020.

\bibitem{zhang2020grannite}
Y.~Zhang, H.~Ren, and B.~Khailany, ``Grannite: Graph neural network inference
  for transferable power estimation,'' in \emph{DAC}.\hskip 1em plus 0.5em
  minus 0.4em\relax IEEE, 2020, pp. 1--6.

\bibitem{xie2021net}
Z.~Xie, R.~Liang, X.~Xu, J.~Hu, Y.~Duan, and Y.~Chen, ``Net2: A graph attention
  network method customized for pre-placement net length estimation,'' in
  \emph{ASP-DAC}.\hskip 1em plus 0.5em minus 0.4em\relax IEEE, 2021, pp.
  671--677.

\bibitem{sutskever2014sequence}
I.~Sutskever, O.~Vinyals, and Q.~V. Le, ``Sequence to sequence learning with
  neural networks,'' 2014.

\bibitem{vaswani2017attention}
A.~Vaswani, N.~Shazeer, N.~Parmar, J.~Uszkoreit, L.~Jones, A.~N. Gomez,
  {\L}.~Kaiser, and I.~Polosukhin, ``Attention is all you need,'' in
  \emph{NIPS}, 2017, pp. 5998--6008.

\bibitem{ITC99}
S.~Davidson, ``Characteristics of the itc’99 benchmark circuits,'' in
  \emph{ITSW}, 1999.

\bibitem{albrecht2005iwls}
C.~Albrecht, ``Iwls 2005 benchmarks,'' in \emph{IWLS}, 2005.

\bibitem{EPFLBenchmarks}
L.~Amar{\'u}, P.-E. Gaillardon, and G.~De~Micheli, ``The epfl combinational
  benchmark suite,'' in \emph{IWLS}, no. CONF, 2015.

\bibitem{takeda2008opencore}
O.~Team, ``Opencores,'' \url{https://opencores.org/}.

\bibitem{selsam2018learning}
D.~Selsam, M.~Lamm, B.~B{\"u}nz, P.~Liang, L.~de~Moura, and D.~L. Dill,
  ``Learning a sat solver from single-bit supervision,'' \emph{arXiv preprint
  arXiv:1802.03685}, 2018.

\bibitem{bommasani2021opportunities}
R.~Bommasani, D.~A. Hudson, E.~Adeli, R.~Altman, S.~Arora, S.~von Arx, M.~S.
  Bernstein, J.~Bohg, A.~Bosselut, E.~Brunskill \emph{et~al.}, ``On the
  opportunities and risks of foundation models,'' \emph{arXiv preprint
  arXiv:2108.07258}, 2021.

\end{thebibliography}










\end{document}